\documentclass[11pt]{article}
\usepackage[in]{fullpage}
\usepackage{enumitem}

\usepackage[hidelinks]{hyperref}
\usepackage[square,numbers]{natbib}

\usepackage{graphicx}
\usepackage{subcaption}

\usepackage{amsmath}
\usepackage{amssymb}
\usepackage{amsthm}
\usepackage{bm}
\usepackage{algorithmic}
\usepackage{algorithm}
\usepackage{mathletters}

\DeclareMathOperator*{\argmax}{argmax}

\newtheorem{definition}{Definition}
\newtheorem{example}{Example}
\newtheorem{fact}{Fact}
\newtheorem{lemma}{Lemma}
\newtheorem{proposition}{Proposition}
\newtheorem{remark}{Remark}
\newtheorem{theorem}{Theorem}

\usepackage{siunitx}



\begin{document}

% \title{Estimating Quality in User-Guided\\Multi-Objective Bandits Optimization}
\title{Estimating Quality in Multi-Objective Bandits Optimization}

\author{Audrey Durand and Christian Gagn\'e \\
Computer Vision and Systems Laboratory \\
% Department of Electrical Engineering and Computer Engineering \\
Universit\'e Laval, Qu\'ebec (QC), Canada\\
\{ audrey.durand.2@ulaval.ca, christian.gagne@gel.ulaval.ca \}
}

\maketitle

\begin{abstract}
    %!TEX root = /Users/audrey/Dropbox/PhD/MOMAB/ArXiv/Latex/paper.tex

Many real-world applications are characterized by a number of conflicting performance measures. As optimizing in a multi-objective setting leads to a set of non-dominated solutions, a preference function is required for selecting the solution with the appropriate trade-off between the objectives. The question is: how good do estimations of these objectives have to be in order for the solution maximizing the preference function to remain unchanged? In this paper, we introduce the concept of preference radius to characterize the robustness of the preference function and provide guidelines for controlling the quality of estimations in the multi-objective setting. More specifically, we provide a general formulation of multi-objective optimization under the bandits setting. We show how the preference radius relates to the optimal gap and we use this concept to provide a theoretical analysis of the Thompson sampling algorithm from multivariate normal priors. We finally present experiments to support the theoretical results and highlight the fact that one cannot simply scalarize multi-objective problems into single-objective problems.

\end{abstract}

%!TEX root = /Users/audrey/Dropbox/PhD/MOMAB/ArXiv/Latex/paper.tex

\section{Introduction}
\label{sec:intro}

Multi-objective optimization (MOO)~\cite{Coello2007} is a topic of great importance for real-world applications. Indeed, optimization problems are characterized by a number of conflicting, even contradictory, performance measures relevant to the task at hand. For example, when deciding on the healthcare treatment to follow for a given sick patient, a trade-off must be made between the efficiency of the treatment to heal the sickness, the side effects of the treatment, and the treatment cost. MOO is often tackled by combining the objective into a single measure (a.k.a.~scalarization). Such approaches are said to be \emph{a priori}, as the preferences over the objectives is defined before carrying out the optimization itself. The challenge lies in the determination of the appropriate scalarization function to use and its parameterization. Another way to conduct MOO consists in learning the optimal trade-offs (the so-called Pareto-optimal set). Once the optimization is completed, techniques from the field of multi-criteria decision-making are applied to help the user to select the final solution from the Pareto-optimal set. These \emph{a posteriori} techniques may require a huge number of evaluations to have a reliable estimation of the objective values over all potential solutions. Indeed, the Pareto-optimal set can be quite large, encompassing a majority, if not all, of the potential solutions. In this work, we tackle the MOO problem where the scalarization function \emph{exists} a priori, but might be unknown, in which case a user can act as a black box for articulating preferences. Integrating the user to the learning loop, she can provide feedback by selecting her preferred choice given a set of options -- the scalarization function lying in her head.

More specifically, we consider problems where outcomes are stochastic and costly to evaluate (e.g., involving a human in the loop). The challenge is therefore to identify the best solutions given random observations sampled from different (unknown) density distributions. We formulate this problem as multi-objective bandits, where we aim at finding the solution that maximizes the preference function while maximizing the performance of the solutions evaluated during the optimization. The Thompson sampling (TS)~\cite{Thompson1933} technique is a typical approach for bandits problems, where potential solutions are tried based on a Bayesian posterior over their expected outcome. Here we consider TS from multivariate normal (MVN) priors for multi-objective bandits.
% Let the \emph{right choice} denote the option that maximize the preference function -- the option that the user would select given that she had knowledge of the Pareto-optimal set. A learning algorithm for the multi-objective bandits setting aims at learning good-enough estimations of the available options to allow the user to make the right choices and its performance depends on the robustness of the preference function to the quality of estimations. We therefore need a measure for characterizing the quality of estimations required in order for the option maximizing the preference function to remain unchanged. For that purpose, we introduce the concept of preference radius providing the tolerance range over objective value estimations, such that the user preference would remain the same as if the Pareto-optimal set was known. We use this concept for providing a theoretical analysis of TS from MVN priors.
We introduce the concept of preference radius providing the tolerance range over objective value estimations, such that the \emph{best option} given the preference function remains unchanged. We use this concept for providing a theoretical analysis of TS from MVN priors.
Finally, we perform some empirical experiments to support the theoretical results and also highlight the importance of tackling multi-objective bandits problems as such instead of scalarizing those under the traditional bandit setting. 

% The original contributions of the paper consist in:
% \begin{itemize}
%     \item providing a general formulation of the MOO under the a priori multi-objective bandits setting;
%     \item proposing the preference radius to characterize the robustness of the preference function to the estimations quality;
%     \item proposing a theoretical analysis of the TS algorithm from MVN priors;
%     \item showing with empirical experiments that multi-objective bandits cannot simply be brought back to single-objective bandits.
% \end{itemize}

%!TEX root = /Users/audrey/Dropbox/PhD/MOMAB/ArXiv/Latex/paper.tex

\section{Multi-Objective Bandits}
\label{sec:momab}

A multi-objective bandits problem is described by a (finite) set of actions $\cA$, also referred to as the \emph{design space}, each of which is associated with a $d$-dimensional expected outcome $\bsmu_a = (\mu_{a, 1}, \dots, \mu_{a, d}) \in \cX \in \Real^d$. For simplicity, we assume that the \emph{objective space} $\cX = [0, 1]^d$. In this episodic game, an agent interacts with an environment characterized by a \emph{preference function}~$f$. The agent iteratively chooses to perform an action $a(t)$ and obtains a noisy observation of $\bsz(t)$.\footnote{Scalars are written unbolded; vectors are boldfaced. The operators $+$, $-$, $\times$, and~$\div$ applied on a vector $\bsv = (v_1, \dots, v_d)$ and a scalar $s$ correspond to the operation between each item of $\bsv$ and $s$, e.g., $\bsv + s = (v_1 + s, \dots, v_d + s)$. These operators applied on two vectors $\bsv = (v_1, \dots, v_d)$ and $\bsu = (u_1, \dots, u_d)$ correspond to itemwise operations between $\bsv$ and $\bsu$, e.g., $\bsv + \bsu = (v_1 + u_1, \dots, v_d + u_d)$.}

An algorithm for a multi-objective bandits problem is a (possibly randomized) method for choosing which action to play next, given a history of previous choices and obtained outcomes, $\cH_t = \{ a(s), \bsz(s) \}_{s = 1}^{t-1}$. Let $\cO = \argmax_{a \in \cA} f(\bsmu_a)$ and let $\star \in \cO$ denote the optimal action. The optimal gap $\Delta_a = f(\bsmu_\star)- f(\bsmu_a)$ measures the expected loss of playing action $a$ instead of the optimal action. The agent's goal is to design an algorithm with low expected (cumulative) regret\footnote{Also known as the scalarized regret~\cite{Drugan2013}.}:
\begin{align}
\label{eqn:regret}
    \kR(T)
    = \sum_{t=1}^T \big( f(\bsmu_\star) - f(\bsmu_{a(t)}) \big)
    = \sum_{t = 1}^T \sum_{a \in \cA} \Pr[a(t) = a] \Delta_a.
\end{align}
This quantity measures the expected performance of the algorithm compared to the expected performance of an optimal algorithm given knowledge of the outcome distributions, i.e., always sampling from the distribution with the expectation maximizing $f$. Typically, we assume that the algorithm maintains one estimate $\bstheta_a(t)$ per action $a$ on time $t$. Let $\cO(t) = \argmax_{a \in \cA} f(\bstheta_a(t))$ denote the set of actions with an estimate maximizing $f$. The algorithm faces a trade-off between playing an action $a(t) \in \cO(t)$ and choosing to gather an additionnal sample from a relatively unexplored action in order to improve its estimate. Alg.~\ref{alg:mobandits} describes this multi-objective bandits problem.

\begin{algorithm}[t]
    On each episode $t \geq 1$:
    \begin{enumerate}[nolistsep]
        \item The agent selects action $a(t)$ to play given $\cO(t)$.
        \item The agent observes $\bsz(t) = \bsmu_{a(t)} + \bsxi(t)$, where $\bsxi(t)$ are i.i.d. random vectors.
        \item The agent updates its estimates.
    \end{enumerate}
    \caption{Multi-objective bandits setting}
\label{alg:mobandits}
\end{algorithm}

In many situations, the environment providing the preference function is a person, let us call her the \emph{expert user}. Unfortunately, people are generally unable to scalarize their choices and preferences. Therefore they cannot explicitely provide their preference function. However, given several options, users can tell which one(s) they prefer (that is $\cO(t)$) and thus can be used as a black box to provide feedback in the learning loop.
% Alg.~\ref{alg:mobandits_user} describes the user-guided multi-objective bandits problem.
%
% \begin{algorithm}[t]
%     On each episode $t \geq 1$:
%     \begin{enumerate}[nolistsep]
%         \item The agent selects action $a(t) \in \cO(t)$ to play.
%         \item The agent observes $\bsz(t) = \bsmu_{a(t)} + \bsxi(t)$, where $\bsxi(t)$ are i.i.d. random variables.
%         \item The agent updates its estimates.
%         \item The agent shows options $\{ \bstheta_a(t) \}_{a \in \cA}$ to the expert user.
%         \item The expert user indicates its preference $\cO(t) = \argmax_{a \in \cA} f(\bstheta_a(t))$.
%     \end{enumerate}
%     \caption{User-guided multi-objective bandits setting}
% \label{alg:mobandits_user}
% \end{algorithm}

\paragraph{Pareto-optimality}

\begin{figure}[t]
    \centering
    \includegraphics[scale=0.75]{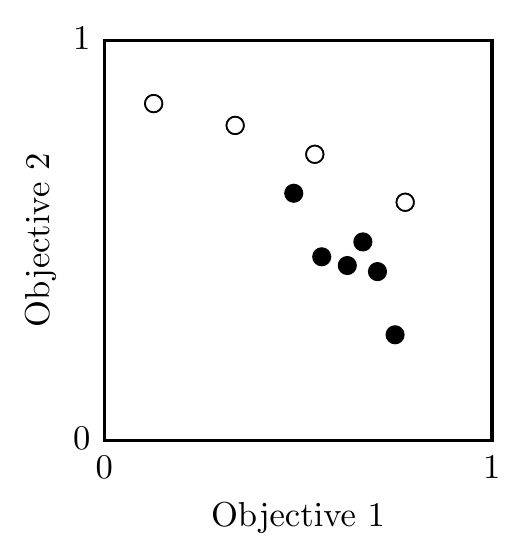}
    \caption{Example of dominated (black) and non-dominated (white) options.}
\label{fig:pareto_front}
\end{figure}

Given two $d$-dimensional options $\bsx = (x_1, \dots, x_d)$ and $\bsy = (y_1, \dots, y_d)$, $\bsx$ is said to dominate, or Pareto-dominate, $\bsy$ (denoted $\bsx \succeq \bsy$) if and only if $x_i > y_i$ for at least one $i$ and $x_i \geq y_i$ otherwise. The dominance is strict (denoted $\bsx \succ \bsy$) if and only if $x_i > y_i$ for all $i = 1, \dots, d$. Finally, the two vectors are incomparable (denoted $\bsx \parallel \bsy$) if $\bsx \nsucc \bsy$ and $\bsy \nsucc \bsx$. Pareto-optimal options represent the best compromises amongst the objectives and are the only options that need to be considered in an application. We say that these options constitute the Pareto front $\cP = \{ a : \nexists \bsmu_b \succeq \bsmu_a \}_{a, b \in \cA}$. Fig.~\ref{fig:pareto_front} shows an example of dominated and non-dominated expected outcomes in a $d = 2$ objectives space. A user facing a multi-criteria decision making problem must select her preferred non-dominated option. Dominated options are obviously discarded by default.

\paragraph{Related Works}

The multi-objective bandits problem has already been addressed in the a posteriori setting, where the goal is to discover the whole Pareto front for a posteriori decision making~\cite{Drugan2013,Yahyaa2015}. This is different from the a priori optimization problem tackled here. The aim of algorithms in the a posteriori setting is to simultaneously minimize the Pareto-regret and the unfairness metrics. Also known as the $\epsilon$-distance~\cite{Laumanns2002}, the Pareto-regret associated with playing action $a$ is the minimum value $\epsilon_a$ such that $\bsmu_a + \epsilon_a$ is not dominated by any other actions. In other words, any action standing on the front is considered equally good by the expert user. This is like considering that $\cO = \cP$, which corresponds to the preference function $f(\bsmu_\star) = 1$, $f(\bsmu_a) = 1 - \epsilon_a$, such that $\Delta_a = \epsilon_a$. Note that any algorithm optimizing a single objective could minimize the Pareto-regret regardless of the other objectives. This is addressed by the unfairness metric, measuring the disparity in the amount of plays of non-dominated actions -- the idea being to force algorithms to explore the whole Pareto front evenly.

In MOO settings~\cite{Zuluaga2013}, the goal is to identify the Pareto-optimal set $\cP$ without evaluating all actions. The quality of a solution $\cS$ is typically given by the hypervolume error $V(\cP) - V(\cS)$, where the $V(\cP)$ is the volume enclosed between the origin and $\{ \bsmu_a \}_{a \in \cP}$ (and similarly for $\cS$). However, the hypervolume error does not give information about the quality of the estimation of actions. Identifying the Pareto front alone does not guarantee that the actions are well estimated and, therefore, that an expert user choice based on these estimations would lead to the right choice.

%!TEX root = /Users/audrey/Dropbox/PhD/MOMAB/ArXiv/Latex/paper.tex

\section{Preference Radius}
\label{sec:preference_radius}

Let $\bstheta_a(t)$ denote the estimation associated with action $a$ on episode $t$ and let $\cP(t) = \{ a : \nexists \bstheta_b(t) \succ \bstheta_a(t) \}_{a, b \in \cA}$ denote the estimated Pareto front given these options. By definition, the optimal options are $\cO(t) \subseteq \cP(t)$. Let
\begin{align*}
    B(\bsc, r) \subseteq \{ \bsx \in \cX : |x_i - c_i| < r, ~ i = 1, \dots, d \}
\end{align*}
denote a ball of center $\bsc$ and radius $r$. In order to characterize the difficulty of a multi-objective bandits setting, we introduce the following quantity.
\begin{definition}
For each action $a \in \cA$, we define the preference radius $\rho_a$ as any radius such that if $\bstheta_a(t) \in B(\bsmu_a, \rho_a)$ for all actions, then
\begin{align*}
    \exists \star \in \cO : \star \in \cO(t) \quad \text{and} \quad a \not\in \cO(t) ~ \forall a \in \cA, a \not\in \cO.
\end{align*}
\end{definition}
The radii correspond to the \emph{robustness} of the preference function, that is to which extent can actions be poorly estimated simultaneously before the set of optimal options changes. The radius $\rho_a$ is directly linked to the gap $\Delta_a = f(\bsmu_\star)- f(\bsmu_a)$. For a suboptimal action, a large radius indicates that this action is far from being optimal. Also, the preference radii of suboptimal actions depend on the preference radius of the optimal action(s). Larger optimal action radii imply smaller radii for suboptimal actions. Note that if all actions estimates stand in their preference balls, being greedy is optimal.

Let $\alpha_1, \dots, \alpha_d \in [0, 1]$ denote \emph{weights} such that $\sum_{i = 1}^d \alpha_i = 1$. The weighted $L_p$ metric $f(\bsx)= \big( \sum_{i = 1}^d \alpha_i x_i^p \big)^{1/p}$ with $p \geq 1$ is often used to represent decision functions. This function is known as the linear scalarization when $p = 1$ and as the Chebyshev scalarization when $p = \infty$. The following examples show the link between the preference radii and the gap for these two common functions.

\begin{figure}[t]
    \centering
    \begin{subfigure}[b]{0.35\textwidth}
        \includegraphics[scale=0.75]{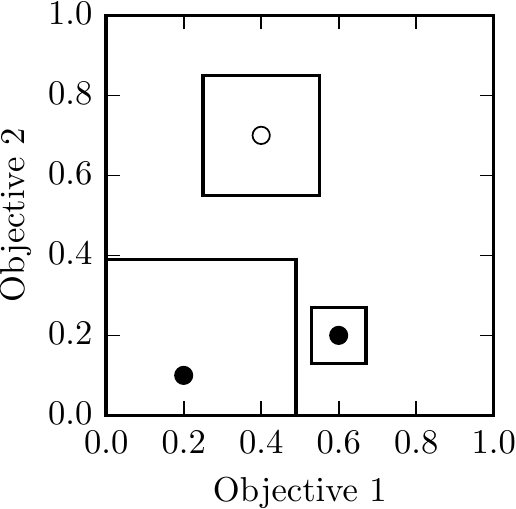}
    \end{subfigure}
    \qquad
    \begin{subfigure}[b]{0.35\textwidth}
        \includegraphics[scale=0.75]{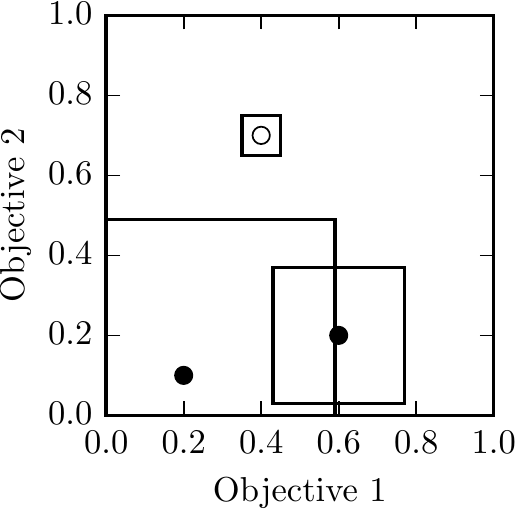}
    \end{subfigure}
    \caption{Examples of preference radii around the optimal (white) and suboptimal (black) actions given the linear preference function $f(\bsx) = 0.4 x_1 + 0.6 x_2$.}
\label{fig:pref_radius:example:weigthed_sum}
\end{figure}

\begin{example}[Linear]
\label{ex:linear}
    The linear scalarization function is given by
    \begin{align*}
        f(\bsx) = \sum_{i = 1}^d \alpha_i x_i.
    \end{align*}
    Consider the optimal action $\star$ and the suboptimal action $a$. By definition of the preference radii, we have that
    \begin{align*}
        \min_{\bstheta_\star \in B(\bsmu_\star, \rho_\star)} f(\bstheta_\star) & > \max_{\bstheta_a \in B(\bsmu_a, \rho_a)} f(\bstheta_a) \\
        \sum_{i = 1}^d (\alpha_i \mu_{\star, i} - \alpha_i \rho_\star) &> \sum_{i = 1}^d (\alpha_i \mu_{a, i} + \alpha_i \rho_a) \\
        f(\bsmu_\star) - \rho_\star & > f(\bsmu_a) + \rho_a \\
        \Delta_a & > \rho_\star + \rho_a.
    \end{align*}
    Fig.~\ref{fig:pref_radius:example:weigthed_sum} shows examples of preference radii with a linear preference function.
%In both cases, as long as each option stays in its associated radius, the expert user preference won't change.
\end{example}

\begin{figure}[t]
    \centering
    \begin{subfigure}[b]{0.35\textwidth}
        \includegraphics[scale=0.75]{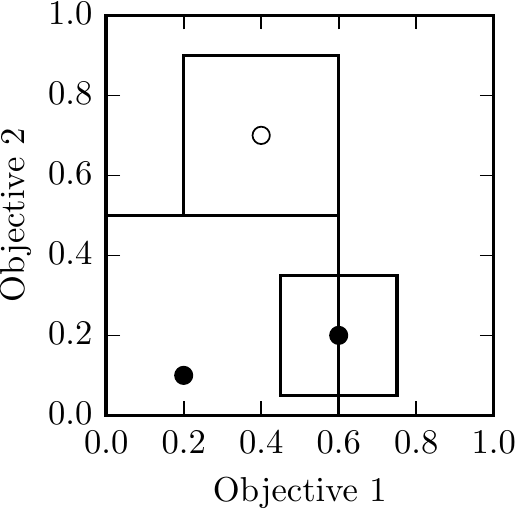}
    \end{subfigure}
    \qquad
    \begin{subfigure}[b]{0.35\textwidth}
        \includegraphics[scale=0.75]{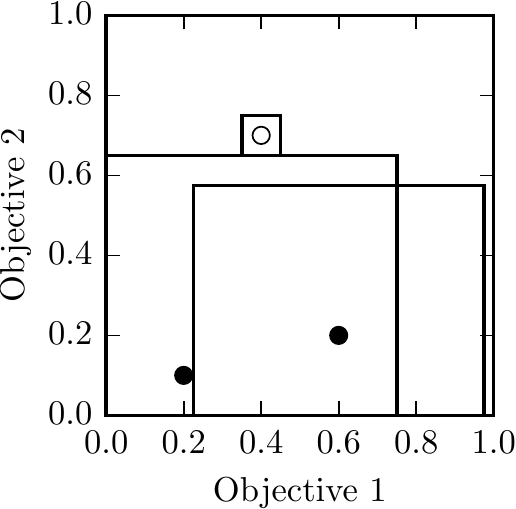}
    \end{subfigure}
    \caption{Examples of preference radii around the optimal action (white) and suboptimal actions (black) given a Chebyshev function with $\alpha_1 = 0.4$ and $\alpha_2 = 0.6$.}
\label{fig:pref_radius:example:chebyshev}
\end{figure}

\begin{example}[Chebyshev]
\label{ex:chebyshev}
    The Chebyshev scalarization~\cite{Bowman1976} function is given by
    \begin{align*}
        f(\bsx) = \max_{1 \leq i \leq d} \alpha_i x_i.
    \end{align*}
    Consider the optimal and suboptimal actions $\star$ and $a$, and let
    \begin{align*}
        i_\star = \argmax_{1 \leq i \leq d} \alpha_i (\mu_{\star, i} - \rho_\star), \quad
        i_a = \argmax_{1 \leq i \leq d} \alpha_i (\mu_{a, i} - \rho_a).
    \end{align*}
    By definition of the preference radii, we have that
    \begin{align*}
        \min_{\bstheta_\star \in B(\bsmu_\star, \rho_\star)} f(\bstheta_\star) & > \max_{\bstheta_a \in B(\bsmu_a, \rho_a)} f(\bstheta_a) \\
        \max_{1 \leq i \leq d} \alpha_i (\mu_{\star, i} + \rho_\star) & > \max_{1\leq i \leq d} \alpha_i (\mu_{a, i} - \rho_a) \\
        \alpha_{i_\star} \mu_{\star, i} - \alpha_{i_\star} \rho_\star & > \alpha_{i_a} \mu_{a, i} + \alpha_{i_a} \rho_a  \\
        f(\bsmu_\star) - \alpha_{i_\star} \rho_\star &> f(\bsmu_a) + \alpha_{i_a} \rho_a \\
        \Delta_a &> \alpha_{i_\star} \rho_\star + \alpha_{i_a} \rho_a.
    \end{align*}
    The difficulty here is that $i_\star$ and $i_a$ respectively depend on $\rho_\star$ and $\rho_a$. Consider a 2-objective setting, we can define
    \begin{align*}
        \tau_\star = \frac{\alpha_2 \mu_{\star, 2} - \alpha_1 \mu_{\star, 1}}{\alpha_2 - \alpha_1}, \quad
        \tau_a = \frac{\alpha_1 \mu_{a, 1} - \alpha_2 \mu_{a, 2}}{\alpha_2 - \alpha_1}
    \end{align*}
    as thresholds such that
    \begin{align*}
        i_\star = \bigg\{
            \begin{array}{ll}
            1 & \quad \text{if} \quad \rho_\star > \tau_\star \\
            2 & \quad \text{otherwise}
            \end{array}, \quad
        i_a = \bigg\{
            \begin{array}{ll}
            1 & \quad \text{if} \quad \rho_a < \tau_a \\
            2 & \quad \text{otherwise}.
            \end{array}
    \end{align*}
    Fig.~\ref{fig:pref_radius:example:chebyshev} shows examples of preference radii with a Chebyshev preference function. 
\end{example}

\begin{figure}[t]
    \centering
    \begin{subfigure}[b]{0.35\textwidth}
        \includegraphics[scale=0.75]{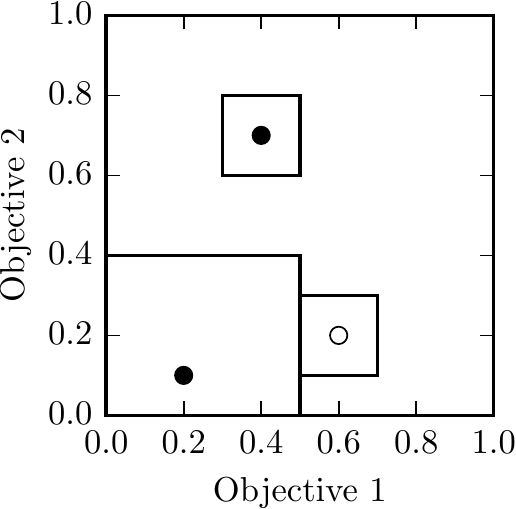}
        \caption{$\ell = 1$, $\epsilon_2 = 0.1$}
    \end{subfigure}
    \qquad
    \begin{subfigure}[b]{0.35\textwidth}
        \includegraphics[scale=0.75]{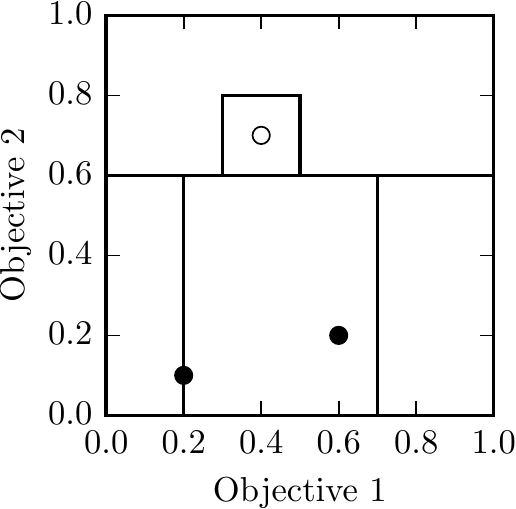}
        \caption{$\ell = 2$, $\epsilon_1 = 0.3$}
    \end{subfigure}
    \caption{Examples of preference radii around the optimal (white) and suboptimal (black) actions given two different configurations of $\epsilon$-contraint.}
\label{fig:pref_radius:example:econstraint}
\end{figure}

Outside $L_p$ metrics, other scalarization functions are often based on constraints. For example, using the $\epsilon$-constraint scalarization technique, a user assigns a constraint to every objective except a target objective $\ell$. All options that fail to respect one of the contraints receive a value of 0, while the options that respect all constraints get a value of $x_\ell$. The following example shows the relation between the preference radius and the gap given a preference function that is articulated as an $\epsilon$-constraint scalarization technique.
\begin{example}[Epsilon-constraint]
\label{ex:epsilon-constraint}
    The $\epsilon$-constraint function is given by
    \begin{align*}
    f(\bsx) = \bigg\{
        \begin{array}{ll}
        x_\ell & \quad \text{if} \quad x_i \geq \epsilon_i \quad \forall i \in \{ 1, \dots, d \}, i \neq \ell \\
        0 & \quad \text{otherwise}.
        \end{array}
    \end{align*}
    Consider the optimal and suboptimal actions $\star$ and $a$. By definition of the preference radii, we have that
    \begin{align*}
        \rho_\star \leq \min_{1 \leq i \leq d, i \neq \ell} \mu_{\star, i} - \epsilon_i.
    \end{align*}
    We decompose $\rho_a = \ubar{\rho}_a + \bar{\rho}_a$ such that
    \begin{align*}
        \ubar{\rho}_a = \min \{ 0, \max_{1 \leq i \leq d, i \neq \ell} \epsilon_i - \mu_{a, i} \}
    \end{align*}
    denotes the radius required in order for action a to respect the constraints, that is to obtain $f(\bsmu_a) > 0$, and $\bar{\rho}_a$ denotes the leftover leading to a gap reduction. Finally, we have that
    \begin{align*}
        \mu_{\star, \ell} - \rho_\star > \mu_{a, \ell} + \ubar{\rho}_a + \bar{\rho_a} \quad \text{and} \quad
        \Delta_a > \rho_\star + \rho_a.
    \end{align*}
    Fig.~\ref{fig:pref_radius:example:econstraint} shows examples of preference radii with $\epsilon$-constraint preference functions.
\end{example}

%!TEX root = /Users/audrey/Dropbox/PhD/MOMAB/ArXiv/Latex/paper.tex

\section{Thompson Sampling}
\label{sec:ts}

% In this section we use the preference radius to analyze the Thompson sampling (TS)~\cite{Thompson1933} algorithm using multivariate normal (MVN) priors. Recall that TS
The Thompson sampling (TS)~\cite{Thompson1933} algorithm maintains a posterior distribution $\pi_a(t)$ on the mean $\bsmu_a$ given a prior and the history of observations $\cH_t$. On each episode $t$, one option $\bstheta_a(t)$ is sampled from each posterior distribution $\pi_a(t)$. The algorithm selects $a(t) \in \cO(t)$. Recall that $\cO(t) = \argmax_{a \in \cA} f(\bstheta_a(t))$. Therefore $\Pr[a(t) = a]$ is proportionnal to the posterior probability that $a$ maximizes the preference function given the history $\cH_t$. Let $N_a(t) = \sum_{s = 1}^{t-1} \ind[a(s) = a]$ denote the number of times action $a$ has been played up to episode $t$. Also let
\begin{align*}
    \hat \bsmu_{a, t} = \frac{\sum_{s = 1: a(s) = a}^{t-1} \bsz(s)}{N_a(t)}
    ~\text{and}~
    \hat \bsSigma_{a, t} = \frac{\sum_{s = 1: a(s) = a}^{t - 1} \big( \bsz(s) - \hat \bsmu_a(t) \big) \big( \bsz(s) - \hat \bsmu_a(t) \big)^\top}{N_a(t) - 1}
\end{align*}
respectively denote the empirical mean and covariance, and let $\bsSigma_0$ and $\bsmu_0$ denote priors. For MVN priors, the posterior over $\bsmu_a$ is given by a MVN distribution $\Normal_d (\tilde \bsmu_a(t), \tilde \bsSigma_a(t))$, where
\begin{align*}
    \tilde \bsSigma_a(t) = \big( \bsSigma_0^{-1} + N_a(t) \bsSigma_a^{-1} \big)^{-1} ~ \text{and} ~
    \tilde \bsmu_a(t) = \tilde \bsSigma_a(t) \big( \bsSigma_0^{-1} \bsmu_0 + N_a(t) \bsSigma_a^{-1} \hat \bsmu_a(t) \big)
\end{align*}
for the known covariance matrix $\bsSigma_a$. Since assuming that $\bsSigma_a$ is known might be unrealistic in practice, one can consider the non-informative covariance $\bsSigma_a = \bsI_d$. With non-informative priors $\bsmu_0 = \bszero_{d \times 1}$ and $\bsSigma_0 = \bsI_d$,\footnote{$\bszero_{d \times 1}$ indicates a $d$-elements column vector and $\bsI_d$ indicates a $d \times d$ identity matrix.} this corresponds to a direct extension of the one-dimensional TS from Gaussian priors~\cite{Agrawal2013}. Alg.~\ref{alg:mvn_ts} shows the resulting TS procedure from MVN priors.

\begin{algorithm}[t]
    \begin{algorithmic}[1]
        \FORALL{episode $t \geq 1$}
            \FORALL{action $a \in \cA$}
                \STATE sample $\bstheta_a(t) = \Normal_d \big( (\bsI_d + N_a(t) \bsI_d)^{-1} N_a(T) \hat \bsmu_a(t), (\bsI_d + N_a(t) \bsI_d)^{-1} \big)$
            \ENDFOR
            \STATE $\cO(t) = \argmax_{a \in \cA} f(\bstheta_a(t))$
            \STATE play $a(t) \in \cO(t)$ and observe $\bsz(t)$
        \ENDFOR
    \end{algorithmic}
    \caption{Thompson sampling from MVN priors}
\label{alg:mvn_ts}
\end{algorithm}

The following proposition provides general regret bounds for TS from MVN priors. The next theorem specializes these regret bounds for three well known preference function families using the relation between preference radii and the gap, as discussed in previous examples.

\begin{proposition}
\label{prop:mvn_ts}
    Assuming $\sigma$-sub-Gaussian noise with $\sigma^2 \leq 1/(4d)$, the expected regret of TS from MVN priors (Alg.~\ref{alg:mvn_ts}) is bounded by
    \begin{align*}
        \kR(T)
        & \leq \sum_{a \in \cA, a \neq \star} \bigg[
        (C(d) + 4d) (1 + \sigma) \Delta_a \frac{\ln(d T \Delta_a^2)}{\rho_\star^2} + \frac{4}{\Delta_a}
        + 2 \Delta_a \frac{\ln(d T \Delta_a^2)}{(\rho_a - r_a)^2} \\
        & \qquad \qquad \quad + 2 \sigma^2 \Delta_a \frac{\ln(d T \Delta_a^2)}{r_a^2} \bigg],
    \end{align*}
    where $\rho_\star$, $\rho_a$ are preference radii, $r_a < \rho_a$, and $C(d)$ is such that $e^{-\frac{\sqrt{i}}{\sqrt{18 \pi d \ln i}^d}}\leq \frac{d}{i^2}$ for $i \geq C(d)$ (see Remark~\ref{remark:const_c}).
\end{proposition}

\begin{theorem}
\label{thm:mvn_ts}
    Assume either a linear (Ex.~\ref{ex:linear}), Chebyshev (Ex.~\ref{ex:chebyshev}), or $\epsilon$-constraint (Ex.~\ref{ex:epsilon-constraint}) preference function. Assuming $\sigma$-sub-Gaussian noise with $\sigma^2 \leq 1/(4d)$, the expected regret of TS from MVN priors (Alg.~\ref{alg:mvn_ts}) is bounded by
    \begin{align*}
        \kR(T) \leq \sum_{a \in \cA, a \neq \star} \bigg[
        (8C(d) + 24d + 18 + 72\sigma^2) (1 + \sigma)^2 \frac{\ln(d T \Delta_a^2)}{\Delta_a} + \frac{4}{\Delta_a} \bigg],
    \end{align*}
    where $C(d)$ is such that $e^{-\frac{\sqrt{i}}{\sqrt{18 \pi d \ln i}^d}} \leq \frac{d}{i^2}$ for $i \geq C(d)$ (see Remark~\ref{remark:const_c}). This regret bound is of order $\cO(\sqrt{dNT}\ln d + \sqrt{dNT \ln N})$, where $N = |\cA|$. More specifically, for $d \leq \ln N$, it is of order $\cO(\sqrt{dNT \ln N})$.
\end{theorem}

\begin{remark}
\label{remark:const_c}
    For $d = 1$ we can take $C(d) = e^{14}$. For $d = 2$ we can take $C(d) = e^{24}$, for $d = 3$ we can take $C(d) = e^{35}$, and so on for any $d \in \Nat$.
\end{remark}

For $d = 1$, the order of the regret bounds given by Theorem~\ref{thm:mvn_ts} match the order of the regret bounds for TS from Gaussian priors in the single-objective bandits setting~\cite{Agrawal2013}, assuming $[0, 1]$-bounded outcomes. However we observe that the noise tolerance decreases linearly with the dimension $d$ of the objective space. This means that the more dimensions we have, the less noise we can bear in order for these bounds to hold, \emph{given the provided analysis}.

%!TEX root = /Users/audrey/Dropbox/PhD/MOMAB/ArXiv/Latex/paper.tex

\section{Theoretical Analysis}
\label{sec:analysis}

In this section we start by proving Prop.~\ref{prop:mvn_ts} that provides a regret bound for TS with MVN priors that is independent from the preference function. Then we use the relations between the gap and the preference radius in three preference function families to obtain Theorem~\ref{thm:mvn_ts}.

\subsection{Proof of Prop.~\ref{prop:mvn_ts}}

The following analysis extends the work for the 1-dimensional setting~\cite{Agrawal2013} to the $d$-dimensional setting. We rewrite Eq.~\ref{eqn:regret} as
\begin{align*}
    \kR(T) = \sum_{a \in \cA, a \neq \star} \Delta_a \sum_{t = 1}^T \Pr[a(t) = a],
\end{align*}
where we control $\sum_{t = 1}^T \Pr[a(t) = a]$. The proof relies on several facts (see Appendix~\ref{app:technical_tools}) that extend Chernoff's inequalities and (anti-)concentration bounds from the $1$-dimensional setting to the $d$-dimensional setting using the concepts of Pareto-domination and preference radius. We introduce the following quantities and events to control the quality of mean estimations and the quality of samples.

\begin{definition}[Quantities $r_a$]
    For each suboptimal action $a$, we choose a quantity $r_a < \rho_a$, where $\rho_a$ is a preference radius. By definition of the preference radii, we have $\bsmu_a \prec \bsmu_a + r_a \prec \bsmu_a + \rho_a$. Recall that $f(\bsx) < f(\bsy)$ if $\bsx \prec \bsy$. Hence we have $f(\bsmu_a) < f(\bsmu_a + r_a) < f(\bsmu_a + \rho_a) < f(\bsmu_\star - \rho_\star)$.
\end{definition}

\begin{definition}[Events $E_a^\mu(t)$, $E_a^\theta(t)$]
    For each suboptimal action $a$, define $E_a^\mu(t)$ as the event that $\hat \bsmu_a(t) \prec \bsmu_a + r_a$, and define $E_a^\theta(t)$ as the event that $\bstheta_a(t) \prec \bsmu_a + \rho_a$. More specifically, they are the event that suboptimal action $a$ is well estimated and well sampled, respectively.
\end{definition}

\begin{definition}[Filtration $\cF_t$]
    Define filtration $\cF_t = \{ a(s), \bsz(s) \}_{s = 1, \dots, t-1 }$.
\end{definition}

For suboptimal action $a$, we decompose
\begin{align*}
    \sum_{t=1}^T \Pr[a(t) = a]
    % & = \sum_{t=1}^T \Pr[a(t) = a] \\
    & = \underbrace{\sum_{t=1}^T \Pr[a(t) = a, E_a^\mu(t), E_a^\theta(t)]}_{\text(A)}
    + \underbrace{\sum_{t=1}^T \Pr[a(t) = a, E_a^\mu(t), \overline{E_a^\theta(t)}]}_{\text(B)}
    + \underbrace{\sum_{t=1}^T \Pr[a(t) = a, \overline{E_a^\mu(t)}]}_{\text(C)}
\end{align*}
and control each part separately. In (A), $a$ is played while being well estimated and well sampled. We control this by bounding poor estimation and poor samples for the optimal action. In (B), $a$ is played while being well estimated but poorly sampled. We control this using Gaussian concentration inequalities. In (C), $a$ is played while being poorly estimated. We control this using Chernoff inequalities. Gathering the following results together
% , we obtain
% \begin{align*}
%     \Delta_a \Esp[N_a(T)]
%     & \leq \Delta_a (2C(d) + 8d) (1 + \sigma)^2 \frac{\ln(d T \Delta_a^2)}{\rho_\star^2} + \frac{4}{\Delta_a}
%     + \Delta_a 2 \frac{\ln(d T \Delta_a^2)}{(\rho_a - r_a)^2} \\
%     & \quad + \Delta_a 2 \sigma^2 \frac{\ln(d T \Delta_a^2)}{r_a^2}
% \end{align*}
% for $\sigma^2 \leq 1/(4d)$, where $C(d)$ is such that $e^{-\frac{\sqrt{i}}{\sqrt{18 \pi d \ln i}^d}} \leq \frac{d}{i^2}$ for $i \geq C(d)$.
and summing over all suboptimal actions, we obtain Prop.~\ref{prop:mvn_ts}.

\subsubsection{Bounding (A)}

By definition of TS, for suboptimal $a$ to be played on episode $t$, we must (at least) have $f(\bstheta_a(t)) \geq f(\bstheta_\star(t))$. By definition of event $E_a^\theta(t)$ and the preference radii, we have $f(\bstheta_a(t)) < f(\bstheta_\star(t))$ if $\bstheta_\star(t) \succ \bsmu_\star - \rho_\star$. Let $\tau_k$ denote the time step at which action $\star$ is selected for the $k^\mathrm{th}$ time for $k \geq 1$, and let $\tau_0 = 0$. Note that for any action $a$, $\tau_k > T$ for $k > N_a(T)$ and $\tau_T \geq T$. Then

\begin{align}
\label{eqn:a}
    (A)
    & = \sum_{t=1}^T \Pr[a(t) = a, E_a^\mu(t), E_a^\theta(t) | \cF_t] \nonumber \\
    & \leq \sum_{t=1}^T \Pr[f(\bstheta_a(t)) > f(\bstheta_\star(t)), E_a^\mu(t), E_a^\theta(t) | \cF_t] \nonumber \\
    & \leq \sum_{t=1}^T \Pr[\bstheta_\star(t) \not\succ \bsmu_\star - \rho_\star | \cF_t] \nonumber \\
    & \leq \sum_{k=0}^{L} \Esp \bigg[ \sum_{t=\tau_k+1}^{\tau_{k+1}} \ind[\bstheta_\star \not\succ \bsmu_\star - \rho_\star | \cF_t] \bigg]
    + \sum_{t=\tau_L+1}^T \Pr[\bstheta_\star(t) \not\succ \bsmu_\star - \rho_\star, N_\star(t) > L | \cF_t].
\end{align}
The second inequality uses the fact that the sampling of $\bstheta_\star(t)$ is independent from the events $E_a^\mu(t)$ and $E_a^\theta(t)$. The last inequality uses the observation that $\Pr[\bstheta_\star(t) \not\succ \bsmu_\star - \rho_\star | \cF_t]$ is fixed given $\cF_t$ and that it changes only when $\pi_\star(t)$ changes, that is only when action $\star$ is played. The first sum counts the number of episodes required before action $\star$ has been played $L$ times. The second counts the number of episodes where $\star$ is badly sampled after having been played $L$ times. We use the following Lemma to control the first summation, see Appendix~\ref{app:proof:counts_before_succ}.

\begin{lemma}[Based on Lemma~6 from \cite{Agrawal2013}]
\label{lem:counts_before_succ}
    Let $\tau_k$ denote the time of the $k^\mathrm{th}$ selection of action $\star$. Then, for any $d \in \Nat$ and $\sigma^2 \leq 1/(4d)$,
    \begin{align*}
        \Esp \bigg[ \sum_{t=\tau_k+1}^{\tau_{k+1}} \Pr[\bstheta_\star(t) \not\succ \bsmu_\star - \rho_\star | \cF_t] \bigg]
        \leq C(d) + 4d,
    \end{align*}
    where $C(d)$ is such that $e^{-\frac{\sqrt{i}}{\sqrt{18 \pi d \ln i}^d}} \leq \frac{d}{i^2}$ for $i \geq C(d)$.
\end{lemma}

Now we bound the second summation in Eq.~\ref{eqn:a} by controlling the probability of poorly sampling $\bstheta_\star(t)$ when $N_\star(t) > L$. Let $E_\star(t)$ denote the event that $\hat \bsmu_\star(t) \succ \bsmu_\star - \sigma \rho_\star / (1 + \sigma)$. Then we have
\begin{align*}
    \Pr[\bstheta_\star(t) \not\succ \bsmu_\star - \rho_\star, N_\star(t) > L | \cF_t]
    & \leq \Pr \Big[ \bstheta_\star(t) \not\succ \hat \bsmu_\star(t) - \frac{\rho_\star}{1 + \sigma}, E_\star(t), N_\star(t) > L | \cF_t \Big] \\
    & \qquad + \Pr[\overline{E_\star(t)}, N_\star(t) > L | \cF_t] \\
    & \leq \Pr \Big[ \bstheta_\star(t) \not\in B \Big( \hat \bsmu_\star(t), \frac{\rho_\star}{1 + \sigma} \Big), E_\star(t), N_\star(t) > L | \cF_t \Big] \\
    & \qquad + \Pr \Big[ \hat \bsmu_\star(t) \not\in B \Big( \bsmu_\star - \frac{\sigma \rho_\star}{1 + \sigma} \Big), N_\star(t) > L | \cF_t \Big] \\
    & \leq \frac{d}{2} e^{-\frac{L \rho_\star^2}{2(1 + \sigma)^2}} + 2 d e^{-\frac{L \rho_\star^2}{2(1 + \sigma)^2}}.
\end{align*}
The last inequality uses Facts~\ref{fac:chernoffd} and~\ref{fac:concentration}. With $L = 2 (1 + \sigma)^2 \frac{\ln(d T \Delta_a^2)}{\rho_\star^2}$ we obtain
\begin{align}
\label{eqn:a:prob_bad_theta_star}
    \Pr[\bstheta_\star(t) \not\succ \bsmu_\star - \rho_\star, N_\star(t) > L | \cF_t]
    \leq \frac{5}{2 T \Delta_a^2}.
\end{align}
We use Lem.~\ref{lem:counts_before_succ} and Eq.~\ref{eqn:a:prob_bad_theta_star} in Eq.~\ref{eqn:a} to obtain
\begin{align*}
    (A) \leq (2C(d) + 8d) (1 + \sigma)^2 \frac{\ln(d T \Delta_a^2)}{\rho_\star^2} + \frac{5}{2 \Delta_a^2}
\end{align*}
for $\sigma^2 \leq 1/(4d)$, where $C(d)$ is such that $e^{-\frac{\sqrt{i}}{\sqrt{18 \pi d \ln i}^d}} \leq \frac{d}{i^2}$ for $i \geq C(d)$.

\subsubsection{Bounding (B)}

We control the probability of badly sampling suboptimal action $a$ given that it has been played at least $L$ times. Recall that filtration $\cF_t$ is such that $E_a^\mu(t)$ holds. To that extent we decompose
\begin{align*}
    (B)
    & = \sum_{t=1}^T \Pr[a(t) = a, \overline{E_a^\theta(t)}, E_a^\mu(t), N_a(t) \leq L | \cF_t]
    + \sum_{t=1}^T \Pr[a(t) = a, \overline{E_a^\theta(t)}, E_a^\mu(t), N_a(t) > L | \cF_t] \\
    & \leq \Esp \bigg[ \sum_{t=1}^T \ind[a(t) = a, N_a(t) \leq L | \cF_t] \bigg]
    + \sum_{t=1}^T \Pr[\bstheta_a(t) \not\prec \bsmu_a + \rho_a, N_a(t) > L | \cF_t] \\
    & \leq L + \sum_{t=1}^T \Pr[\bstheta_a(t) \not\prec \hat \bsmu_a(t) + (\rho_a - r_a), N_a(t) > L | \cF_t] \\
    & \leq L + T \frac{d}{2} e^{-\frac{L(\rho_a - r_a)^2}{2}}.
\end{align*}
The first inequality uses the observation that $\Pr[a(t) = a | \cF_t]$ is fixed given $\cF_t$ and the definition of event $\overline{E_a^\theta(t)}$. The second inequality uses the fact that event $E_a^\mu(t)$ holds. The last inequality uses Fact~\ref{fac:concentration}.
% with $\sigma^2 = \frac{1}{N_a(t)+1} \leq \frac{1}{L}$.
With $L = 2 \frac{\ln(d T \Delta_a^2)}{(\rho_a - r_a)^2}$ we obtain
\begin{align*}
    (B) \leq 2 \frac{\ln(d T \Delta_a^2)}{(\rho_a - r_a)^2} + \frac{1}{2 \Delta_a^2}.
\end{align*}

\subsubsection{Bounding (C)}

Similarly to what has been done previously with (B), we can control the probability of badly estimating suboptimal action $a$ given that it has been played at least $L$ times. Then we have
\begin{align*}
    (C)
    & \leq \sum_{t=1}^{T} \Pr[a(t) = a, \overline{E_a^\mu(T)}, N_a(t) \leq L | \cF_t]
    + \sum_{t=1}^T \Pr[a(t) = a, \overline{E_a^\mu(T)}, N_a(T) > L | \cF_t] \\
    & \leq \Esp \bigg[ \sum_{t=1}^T \ind[a(t) = a, N_a(t) \leq L] \bigg] + \sum_{t=1}^T \Pr[\overline{E_a^\mu(T)}, N_a(T) \geq L] \\
    & \leq L + T d e^{-\frac{L r_a^2}{2\sigma^2}}.
\end{align*}
The second inequality uses the observation that $\Pr[a(t) = a | \cF_t]$ is fixed given $\cF_t$. The last inequality uses Fact~\ref{fac:chernoffd}. With $L = 2 \sigma^2 \frac{\ln(d T \Delta_a^2)}{r_a^2}$ we obtain
\begin{align*}
    (C) \leq 2 \sigma^2 \frac{\ln(d T \Delta_a^2)}{r_a^2} + \frac{1}{\Delta_a^2}.
\end{align*}

\subsection{Proof of Theorem~\ref{thm:mvn_ts}}

By definition of the preference radii, given a linear (Ex.~\ref{ex:linear}), Chebyshev (Ex.~\ref{ex:chebyshev}), or $\epsilon$-constraint preference function (Ex.~\ref{ex:epsilon-constraint}), one can take $\rho_\star = \rho_a = \frac{\Delta_a}{2}$, $r_a = \frac{\Delta_a}{6}$. Using these values in Prop.~\ref{prop:mvn_ts}, we obtain Theorem~\ref{thm:mvn_ts}:
\begin{align*}
    \kR(T)
    & \leq \sum_{a \in \cA, a \neq \star} \bigg[
    (8C(d) + 24d + 18 + 72\sigma^2) (1 + \sigma)^2 \frac{\ln(d T \Delta_a^2)}{\Delta_a} + \frac{4}{\Delta_a} \bigg].
\end{align*}
Let $\Delta_a = \delta_a \sqrt{\frac{dN \ln N}{T}}$, for $\delta_a \in (0, \sqrt{\frac{T}{d N \ln N}}]$. The regret is bounded by
\begin{align*}
    \kR(T)
    & \leq (8C(d) + 24d + 18 + 72 \sigma^2) (1 + \sigma)^2  \frac{\sqrt{N T} \ln(d^2 N \ln N)}{\delta_a \sqrt{d \ln N}} + \frac{4 \sqrt{N T}}{\delta_a \sqrt{d \ln N}}
\end{align*}
with $\sigma^2 \leq 1/(4d)$, that is of order $\cO(\sqrt{dNT}\ln d + \sqrt{dNT \ln N})$. More specifically, for $d \leq \ln N$, the regret bound is of order $\cO(\sqrt{dNT \ln N})$.

%!TEX root = /Users/audrey/Dropbox/PhD/MOMAB/ArXiv/Latex/paper.tex

\section{Experiments}
\label{sec:experiments}

Given that the preference function is known a priori, one might be tempted to formalize the problem under the \emph{traditional}, single-objective, bandits setting. This would correspond to optimizing over the expected value of the preference function, $\Esp[f(\bsz(t)) | a(t) =a]$, instead of $f(\bsmu_a)$. In the following experiments, we compare the performance of the TS algorithm from MVN priors (Alg.~\ref{alg:mvn_ts}) in the multi-objective bandits scheme (Alg.~\ref{alg:mobandits}) with the one-dimensional TS from Gaussian priors~\cite{Agrawal2013} applied to the multi-objective bandits problem formalized under the traditional bandits setting (Alg.~\ref{alg:gaussian_ts}).

\begin{algorithm}[t]
    \begin{algorithmic}[1]
        \FORALL{episode $t \geq 1$}
            \FORALL{action $a \in \cA$}
                \STATE sample $\theta_a(t) = \Normal \big( \frac{N_a(T) \hat \mu_a(t)}{N_a(t) + 1}, \frac{1}{N_a(t) + 1} \big)$
            \ENDFOR
            \STATE play $a(t) = \argmax_{a \in \cA} \theta_a(t)$ and observe $f(\bsz(t))$
        \ENDFOR
    \end{algorithmic}
    \caption{Thompson sampling from Gaussian priors~\cite{Agrawal2013}}
\label{alg:gaussian_ts}
\end{algorithm}

\begin{figure}[t]
    \centering
    \begin{subfigure}[b]{0.41\textwidth}
        \centering
        \includegraphics[scale=0.75]{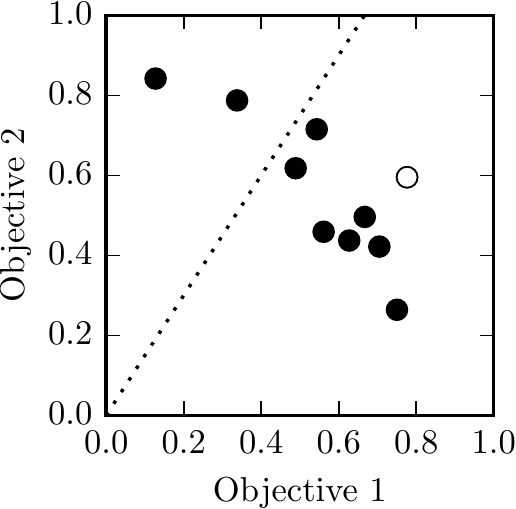}
        \caption{Linear with $\alpha_1 = 0.4$, $\alpha_2 = 0.6$}
        \label{fig:experiments:setting:linear}
    \end{subfigure}
    \qquad
    \begin{subfigure}[b]{0.41\textwidth}
        \centering
        \includegraphics[scale=0.75]{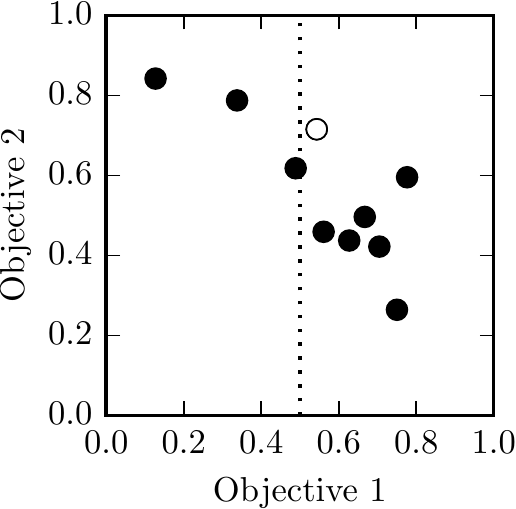}
        \caption{$\epsilon$-constraint with $\ell = 2$, $\epsilon_1 = 0.5$}
        \label{fig:experiments:setting:econstraint}
    \end{subfigure}
    \caption{Expected outcomes for optimal (white) and suboptimal (black) actions. The dotted line shows the preference function (left) and the $\epsilon_1$ constraint (right).}
\label{fig:experiments:setting}
\end{figure}

We randomly generate a 10-action setting with $d = 2$ objectives, such that the objective space is $\cX = [0, 1]^2$. We consider settings where outcomes are sampled from multivariate normal distributions with covariance
$
    \bsSigma_a =
        \begin{bmatrix}
            0.10 & 0.05 \\
            0.05 & 0.10
        \end{bmatrix}
$ for all $a \in \cA$ and from multi-Bernoulli distributions. A sample $\bsz \sim \Bern_d(\bsmu)$ from a $d$-dimensional multi-Bernoulli distribution with mean $\bsmu$ is such that $z_i \sim \Bern(\mu_i)$. Experiments are conducted using the linear preference function
\begin{align*}
    f(\bsx) = 0.4 x_1 + 0.6 x_2, \quad \bsx \in \cX,
\end{align*}
and the $\epsilon$-constraint preference function
\begin{align*}
    f(\bsx) = \bigg\{
        \begin{array}{ll}
            x_2 & \quad \text{if} \quad x_1 \geq 0.5 \\
            0 & \quad \text{otherwise}
        \end{array}, \quad \bsx \in \cX.
\end{align*}
Tab.~\ref{tab:experiments:setting} gives the expected outcomes for all actions along with the associated preference value and gap given the preference function. Fig.~\ref{fig:experiments:setting} shows the expected outcomes and illustrates the preference function. We observe that the optimal action is different for the two preference functions. Each experiment is conducted over \num[group-separator={,}]{10000} episodes and repeated 100 times. Repetitions have been made such that the noise $\bsxi(t)$ is the same for all tested approaches on \emph{the same repetition}. Therefore we can compare the performance of different approaches on the same repetition. The goal is to minimize the cumulative regret (Eq.~\ref{eqn:regret}).

\begin{table}[t]
    \centering
    \caption{Expected outcomes with preference values and gap for both preference functions. The expected outcome for the optimal action is shown in bold.}
    \begin{tabular}{ccccc}
        \hline
        $\bsmu_a$ & \multicolumn{2}{c}{$f(\bsmu_a)$} & \multicolumn{2}{c}{$\Delta_a$} \\
        & Linear & $\epsilon$-constraint & Linear & $\epsilon$-constraint \\
        \hline
        $(0.56, 0.46)$ & 0.50 & 0.46 & 0.17 & 0.26 \\
        $(0.75, 0.26)$ & 0.46 & 0.26 & 0.21 & 0.46 \\
        $(0.34, 0.79)$ & 0.61 & 0.00 & 0.06 & 0.72 \\
        $(0.67, 0.50)$ & 0.56 & 0.50 & 0.11 & 0.22 \\
        $(0.70, 0.42)$ & 0.54 & 0.42 & 0.13 & 0.29 \\
        $(0.54, 0.72)$ & 0.65 & \textbf{0.72} & 0.02 & 0.00 \\
        $(0.49, 0.62)$ & 0.57 & 0.00 & 0.10 & 0.72 \\
        $(0.13, 0.84)$ & 0.56 & 0.00 & 0.11 & 0.72 \\
        $(0.78, 0.60)$ & \textbf{0.67} & 0.60 & 0.00 & 0.12 \\
        $(0.63, 0.44)$ & 0.51 & 0.44 & 0.16 & 0.28 \\
        \hline
    \end{tabular}
\label{tab:experiments:setting}
\end{table}

\begin{figure}[t]
    \centering
    \begin{subfigure}[b]{0.49\textwidth}
        \centering
        \includegraphics[scale=0.75]{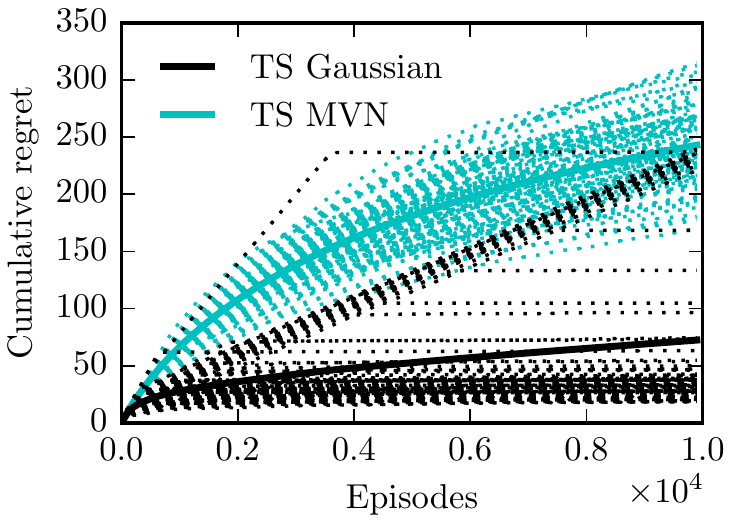}
        \caption{Multi-Bernoulli, linear}
    \end{subfigure}
    \begin{subfigure}[b]{0.49\textwidth}
        \centering
        \includegraphics[scale=0.75]{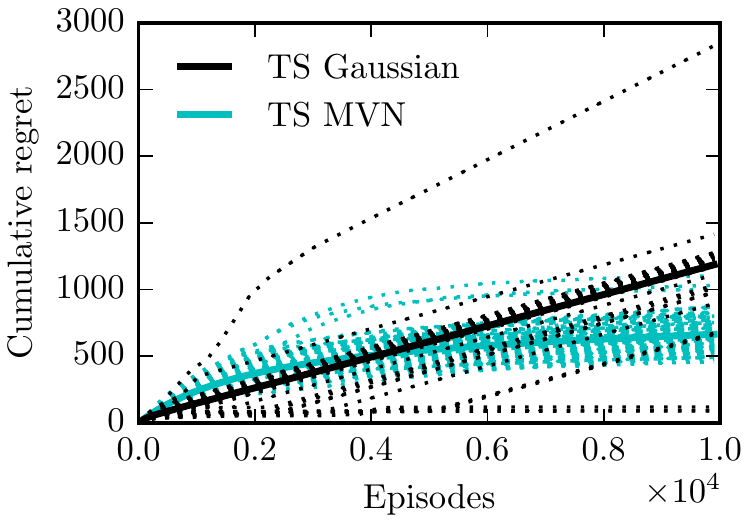}
        \caption{Multi-Bernoulli, $\epsilon$-constraint}
    \end{subfigure}

    \begin{subfigure}[b]{0.49\textwidth}
        \centering
        \includegraphics[scale=0.75]{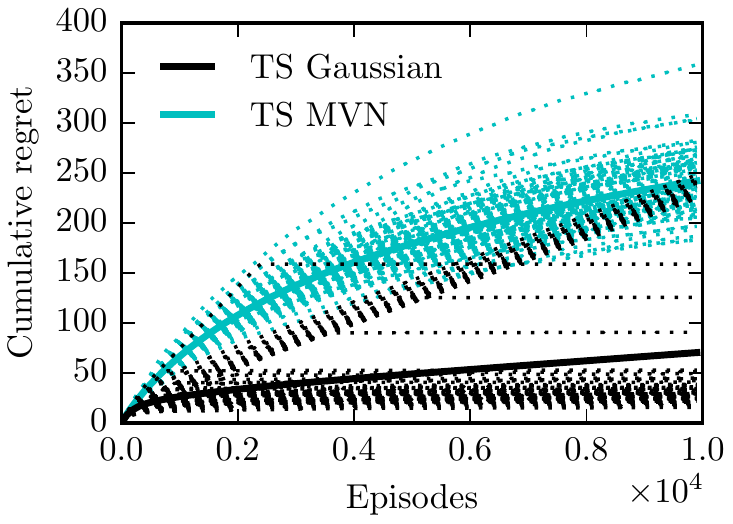}
        \caption{Multivariate normal, linear}
    \end{subfigure}
    \begin{subfigure}[b]{0.49\textwidth}
        \centering
        \includegraphics[scale=0.75]{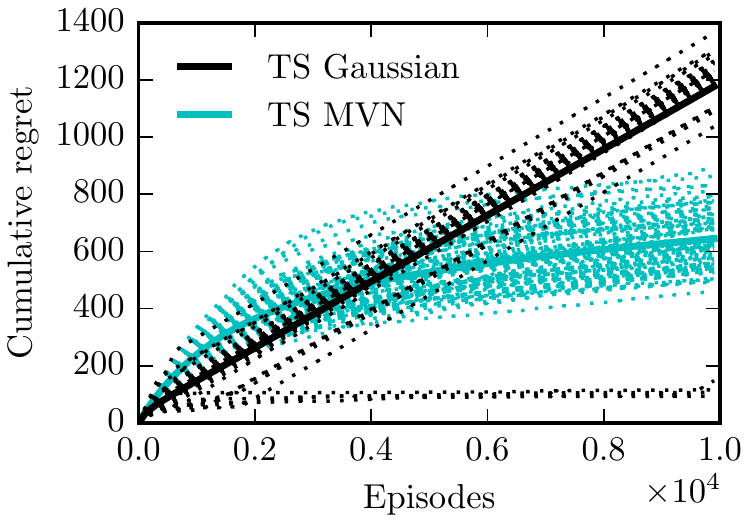}
        \caption{Multivariate normal, $\epsilon$-constraint}
    \end{subfigure}
    \caption{Cumulative regret over episodes for tested outcome distributions and preference functions. Fat lines indicate the average over repetions and dotted lines indicate each individual repetition.}
\label{fig:experiments:results}
\end{figure}

Fig.~\ref{fig:experiments:results} shows the cumulative regret of TS from MVN priors and TS from Gaussian priors (in the traditional bandits formulation) for both outcome distributions and preference functions. We observe that the cumulative regret growth rate for TS from MVN priors appears to match the order of the provided theoretical bounds (Theorem~\ref{thm:mvn_ts}). Results also show that, though it might be appealing to address a multi-objective problem as a single-objective bandits problem, it is not a good idea. Consider the $\epsilon$-constraint preference function used in this experiment. It is evaluated as 0 if $z_1(t) < 0.5$, otherwise to $z_2(t)$. With multi-Bernoulli outcomes, for example, this means that $\Pr[f(\bsz(t)) = 1] = \mu_{a(t), 1} \mu_{a(t), 2}$. Given that, $\argmax_{a \in \cA} f(\bsmu_a) \neq \argmax_{a \in \cA} \Esp[f(\bsz(t)) | a(t) =a]$. Since the action considered as optimal in the single-objective formulation is not the same as the optimal action in the multi-objective problem, TS with Gaussian priors converges to the \emph{wrong} action, hence the linear regret.

%!TEX root = /Users/audrey/Dropbox/PhD/MOMAB/ArXiv/Latex/paper.tex

\section{Conclusion}
\label{sec:conclusion}

In this work, we have addressed the online multi-objective optimization problem under the multi-objective bandits setting. Unlike previous formulations, we work in the a priori setting, where there exists a preference function to be maximized. However, acting in the the proposed setting would not require the preference function to be \emph{known}. Indeed, it would be sufficient for an expert user to pick her preferred estimate among a set of options with no requirement of providing an actual, real valued, evaluation of each option. We have introduced the concept of preference radius to characterize the difficulty of a multi-objective setting through the robustness of the preference function to the quality of estimations available. We have shown how this measure relates to the gap between the optimal action and the recommended action by a learning algorithm. We have used this new concept to provide a theoretical analysis of the Thompson sampling algorithm from multivariate normal priors in the multi-objective setting. More specifically, we were able to provide regret bounds for three families of preference functions. Empirical experiments confirmed the expected behavior of the multi-objective Thompson sampling in terms of cumulative regret growth. Results also highlight the important fact that one cannot simply reduce a multi-objective setting to a traditional, single-objective, setting since this might cause a change in the optimal action. Future work includes the application of the proposed approach to a real world application.

\subsubsection*{Acknowledgements}

This work was supported through funding from NSERC (Canada). We also thank Julien-Charles L\'evesque for insightful comments and Annette Schwerdtfeger for proofreading.

\bibliographystyle{plainnat}
%\bibliography{references,paper}
\bibliography{references}

\newpage
\appendix
{\LARGE\textbf{Appendix}}
%!TEX root = /Users/audrey/Dropbox/PhD/MOMAB/ArXiv/Latex/paper.tex

\section{Technical Tools}
\label{app:technical_tools}

\begin{fact}[$d$-dimensional Chernoff]
\label{fac:chernoffd}
    Let $X_1, \dots, X_N$ be i.i.d. $\sigma$-sub-Gaussian variables with values in such that $\Esp[X] = \mu$. Let $\hat \mu_N = \frac{1}{N} \sum_{i=1}^N X_i$. Then, as shown by \cite{Rigollet2015}, for any $a \geq 0$,
    \begin{align*}
        \Pr[|\hat \mu_N - \mu| \geq a] \leq 2 e^{-\frac{N a^2}{2 \sigma^2}}.
    \end{align*}
    Now consider the the multivariate setting where $\bsX_1, \dots, \bsX_N$ are i.i.d. $d$-dimensional $\sigma$-sub-Gaussian variables such that $\Esp[\bsX] = \bsmu$ and $\hat \bsmu_N = \frac{1}{N} \sum_{i=1}^N \bsX_i$. Then for any $a \geq 0$,
    \begin{align*}
        \Pr[\hat \bsmu_N \succeq \bsmu + a]
        & = \Pr[(\hat \mu_{N, 1} \geq \mu_1 + a) \wedge \dots \wedge (\hat \mu_{N, d} \geq \mu_d + a)]
        \leq e^{-\frac{d N a^2}{2 \sigma^2}}, \\
        \Pr[\hat \bsmu_N \not\preceq \bsmu + a]
        & \leq \Pr[(\hat \mu_{N, 1} \geq \mu_1 + a) \vee \dots \vee (\hat \mu_{N, d} \geq \mu_d + a)]
        \leq d e^{-\frac{N a^2}{2 \sigma^2}}, \\
        \Pr[\hat \bsmu_N \not\in B(\bsmu, a)]
        & \leq \Pr[(|\hat \mu_{N, 1} - \mu_1| \geq a) \vee \dots \vee (|\hat \mu_{N, d} - \mu_d| \geq a)]
        \leq 2 d e^{-\frac{N a^2}{2 \sigma^2}}.
    \end{align*}
\end{fact}

\begin{fact}[$d$-dimensional Gaussian concentration]
\label{fac:concentration}
    Let $X$ be a Gaussian random variable with mean $\mu$ and standard deviation $\sigma$. The following concentration is derived~\cite{Agrawal2013} from \cite{Abramowitz1964} for $z \geq 1$:
    \begin{align*}
        \Pr[|X - \mu| > z \sigma] \leq \frac{1}{2} e^{-z^2/2}.
    \end{align*}
    Now consider the multivariate setting where $\bsX$ denotes a $d$-dimensional Gaussian random variable with mean $\bsmu$ and diagonal covariance $\bsSigma$. Then for $z \geq 1$,
    \begin{align*}
        \Pr[\bsX \succ \bsmu + z \sqrt{\diag(\bsSigma)}]
        & = \Pr[(X_1 > \mu_1 + z \sigma_1) \wedge \dots \wedge (X_d > \mu_d + z \sigma_d)]
        \leq \bigg( \frac{1}{4} e^{-z^2/2} \bigg)^d, \\
        \Pr[\bsX \not\prec \bsmu + z \sqrt{\diag(\bsSigma)}]
        & \leq \Pr[(X_1 \geq \mu_1 + z \sigma_1) \vee \dots \vee (X_d \geq \mu_d + z \sigma_d)]
        \leq \frac{d}{4} e^{-z^2/2}, \\
        \Pr[\bsX \not\in B(\bsmu, z \sqrt{\diag(\bsSigma)})]
        & \leq \Pr[(|X_1 - \mu_1| \geq z \sigma_1) \vee \dots \vee (|X_d - \mu_d| \geq z \sigma_d)]
        \leq \frac{d}{2} e^{-z^2/2}.
    \end{align*}
\end{fact}

\begin{fact}[$d$-dimensional Gaussian anti-concentration]
\label{fac:anti_concentration}
    Let $X$ be a Gaussian random variable with mean $\mu$ and standard deviation $\sigma$. The following concentration is derived~\cite{Agrawal2013} from \cite{Abramowitz1964} for $z \geq 1$:
    \begin{align*}
        \Pr[X > \mu + z \sigma] \geq \frac{z}{\sqrt{2 \pi} (z^2 + 1)} e^{-z^2/2}.
    \end{align*}
    Now consider the multivariate setting where $\bsX$ denotes a $d$-dimensional Gaussian random variable with mean $\bsmu$ and diagonal covariance $\bsSigma$. Then for $z \geq 1$,
    \begin{align*}
        \Pr[\bsX \succ \bsmu + z \sqrt{\diag(\bsSigma)}]
        & = \Pr[(X_1 > \mu_1 + z \sigma_1) \wedge \dots \wedge (X_d > \mu_d + z \sigma_d)]
        \geq \bigg( \frac{z}{\sqrt{2 \pi} (z^2 + 1)} e^{-z^2/2} \bigg)^d.
    \end{align*}
\end{fact}

\section{Proof of Lemma~\ref{lem:counts_before_succ}}
\label{app:proof:counts_before_succ}

\begin{proof}
    Let $\Theta_j$ denote a $\Normal_d (\hat \bsmu_\star(\tau_j+1), (I_d + N_\star(\tau_j+1)I_d)^{-1})$ distributed multivariate normal random variable. Let $G_j$ be a geometric variable denoting the number of consecutive independent trials until $\Theta_j \succ \bsmu_\star - \rho_\star$. Then observe that 
    \begin{align*}
         \Esp \bigg[ \sum_{t=\tau_k+1}^{\tau_{k+1}} \Pr[\bstheta_\star(t) \not\succ \bsmu_\star - \rho_\star | \cF_t] \bigg]
         \leq \Esp[G_j]
         = \sum_{i = 1}^\infty \Pr[G_j \geq i].
    \end{align*}
    We want to bound the expected value of $G_j$ by a constant for all $j$. Consider any integer $i \geq 1$, let $z = \sqrt{\ln i^{1/d}}$, and let $\mathrm{MAX}_i$ denote the \emph{maximum preference} of $i$ independent samples of $\Theta_j$, that is $\max_{1 \leq i \leq j} f(\Theta_j)$. We abbreviate $\hat \bsmu_\star(\tau_j+1)$ as $\hat \bsmu_\star$ and $N_\star(\tau_j+1)$ as $N_\star$ in the following. Then
    \begin{align*}
        \Pr[G_j < i]
        & \geq \Pr[\mathrm{MAX}_i \succ \bsmu_\star - \rho_\star] \\
        & \geq \Pr \Big[ \mathrm{MAX}_i \succ \hat \bsmu_\star + \frac{z}{\sqrt{N_\star}} \Big| \hat \bsmu_\star + \frac{z}{\sqrt{N_\star}} \succeq \bsmu_\star - \rho_\star \Big]
        \cdot \Pr \Big[ \hat \bsmu_\star + \frac{z}{\sqrt{N_\star}} \succeq \bsmu_\star - \rho_\star \Big].
    \end{align*}
    Using Fact~\ref{fac:anti_concentration}, this gives
    \begin{align*}
        \Pr \Big[ \mathrm{MAX}_i \succ \hat \bsmu_\star + \frac{z}{\sqrt{N_\star}} \Big| \hat \bsmu_\star + \frac{z}{\sqrt{N_\star}} \succeq \bsmu_\star - \rho_\star \Big]
        & \geq 1 - \Bigg( 1 - \bigg( \frac{1}{\sqrt{2\pi}} \frac{z}{z^2 + 1} e^{-z^2/2} \bigg)^d \Bigg)^i \\
        & = 1 - \Bigg( 1 - \bigg( \frac{1}{\sqrt{2\pi}} \frac{\sqrt{\ln i^{1/d}}}{(\ln i^{1/d} + 1)} \frac{1}{\sqrt{i^{1/d}}} \bigg)^d \Bigg)^i \\
        & \geq 1 - \Bigg( 1 - \bigg( \frac{1}{\sqrt{18 \pi d i^{1/d} \ln i}} \bigg)^d \Bigg)^i \\
        & \geq 1 - e^{-\frac{\sqrt{i}}{\sqrt{18 \pi d \ln i}^d}},
    \end{align*}
    where the second inequality uses that $\ln i^{1/d} + 1 < 3 \ln i$ and the last inequality uses that $1 - x < e^{-x}$. Also, using Fact~\ref{fac:chernoffd}, we have
    \begin{align*}
        \Pr[\hat \bsmu_\star \succeq \bsmu_\star - \frac{z}{\sqrt{N_\star}}]
        \geq 1 - d e^{-\frac{z^2}{2\sigma^2}}
        = 1 - \frac{d}{i^{1/(2d\sigma^2)}}.
    \end{align*}
    Substituting, we obtain
    \begin{align*}
        \Pr[G_j < i]
        \geq \Big( 1 - e^{-\frac{\sqrt{i}}{\sqrt{4 \pi \ln i}^d}} \Big) \cdot \Big( 1 - \frac{d}{i^{1/(2d\sigma^2)}} \Big)
        \geq 1 - \frac{d}{i^{1/(2d\sigma^2)}} - e^{-\frac{\sqrt{i}}{\sqrt{18 \pi d \ln i}^d}}
    \end{align*}
    and
    \begin{align*}
        \Esp[G_j]
        & = \sum_{i \geq 1} (1 - \Pr[G_j < i]) \\
        & \leq \sum_{i \geq 1} \Big( \frac{d}{i^{1/(2d\sigma^2)}} + e^{-\frac{\sqrt{i}}{\sqrt{18 \pi d \ln i}^d}} \Big) \\
        & \leq C(d) + 2 d \sum_{i \geq 1} \frac{1}{i^{1/(2d\sigma^2)}},
    \end{align*}
    where $C(d)$ is such that $e^{-\frac{\sqrt{i}}{\sqrt{18 \pi d \ln i}^d}} \leq \frac{d}{i^{1/(2d\sigma^2)}}$ for $i \geq C(d)$. We observe that $\sigma^2 \leq 1/(4d)$ is required in order for the sum to converge.
\end{proof}

\end{document}